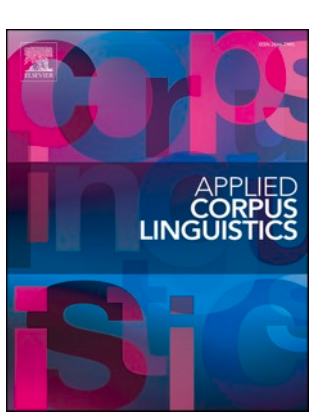

# ChatGPT for President! Presupposed content in politicians versus GPT-generated texts

Davide Garassino [a,1,*], Nicola Brocca [b,2], Viviana Masia [c,3]

[a] Zurich University of Applied Sciences, Department of Applied Linguistics, Theaterstr. 15c, CH-8401, Winterthur, Switzerland
[b] Universität Innsbruck, Institut für Fachdidaktik, Innrain 52, ZiNr. 40513, 5. Stock, Geiwi-Turm. AT-6020, Innsbruck, Austria
[c] Roma Tre University, Department of Foreign Languages, Literatures and Cultures, Via del Valco di San Paolo 19, IT-00142, Rome, Italy

ABSTRACT

This study examines GPT-4′s capability to replicate linguistic strategies used in political discourse, focusing on its potential for manipulative language generation. As Large Language Models (LLMs) become increasingly popular for text generation, concerns have grown regarding their role in spreading fake news and propaganda. This research compares French and Italian political speeches with those generated by GPT-4, with an emphasis on presuppositions – a rhetorical device that may subtly influence audiences by packaging some content as already known at the moment of utterance. Through a corpus-based pragmatic analysis, this study assesses how well GPT-4 can mimic this persuasive strategy. Our findings show that, despite some apparent similarities, a closer look reveals important differences in their distribution and function compared to politicians. For example, GPT-generated texts often rely on change-of-state verbs used in fixed phrases, whereas politicians use presupposition triggers in more varied ways. Such differences, however, are challenging to detect with a 'naked eye,' and this represents a potential risk of LLMs in political and public discourse.

## 1. Introduction

In this study, we present the results of a comparative analysis between political speeches produced by politicians and by GPT-4 (OpenAI, 2024). Our main goal is to assess the extent to which a Large Language Model (LLM) can replicate political discourse and mimic its linguistic features, with particular regard to implicit communication, i.e., a set of strategies by which certain contents can only be reconstructed in the interpretation process.

In particular, we focus on the analysis of presuppositions, given their effectiveness as a subtle persuasive strategy (Sbisà, 2007). By presenting information as already taken for granted and shared in the *common ground*, presuppositions might influence audiences without overtly presenting argumentative claims, which makes them a powerful and potentially manipulative tool in political rhetoric. Our aim is to investigate whether GPT is able to replicate in a 'natural' fashion the use of presuppositions, which may shed further light on its pragmatic competence and its capability of serving persuasive or even manipulative purposes.

By relying on a corpus-based analysis, this study serves both as a confirmatory follow-up investigation of previous work conducted using GPT-3.5 (Garassino et al., 2024) and as a hypothesis-generating exploration. Previous findings pointed to a widespread use of two linguistic devices which are responsible for the conveyance of presupposed meanings, namely, *change-of-state verbs* and *definite descriptions*. Notably, Garassino et al. (2024) found that GPT-3.5 employed change-of-state verbs more frequently than humans when pursuing the same persuasive aim. Moving from these trends, we extended our exploration on this formal feature with a more fine-grained analysis (presented in § 5.2), which allows gaining further insight into GPT-4 underlying architecture and language.

The paper is structured as follows: § 2 outlines the theoretical framework for the analysis and discusses LLMs' text-generating capacities (§ 2.1), with a particular focus on manipulative texts (§ 2.2), as well as their pragmatic competence (§ 2.3). § 2.4 also presents a working definition of presupposition (§ 2.4.1) as well as its potentially manipulative use (§ 2.4.2) and functions (§ 2.4.3) in political discourse. In § 3 the research questions are set out and discussed. § 4 details the corpus composition and the methodological approach (§ 4.1), including the procedures for assessing interrater reliability (§ 4.2). In § 5 the main empirical findings of this paper are discussed. In light of the confirmation of previous findings, the analysis in § 5.1 identifies new hypotheses, which are further investigated in § 5.2. Finally, §§ 6 and 7 discuss the results and consider their broader implications.

* Corresponding author.
*E-mail addresses:* davide.garassino@zhaw.ch (D. Garassino), nicola.brocca@uibk.ac.at (N. Brocca), viviana.masia@uniroma3.it (V. Masia).

[1] Davide Garassino – §§2.1, 2.3, 2.4.2, 4.2, 5.1, 7.
[2] Nicola Brocca – §§1, 2.2, 3, 4.1, 5.2, 6.
[3] Viviana Masia – §§2.4, 2.4.1, 2.4.3.

## 2. State of the art

This study lies at the intersection of recent research strands on LLMs that focus on their ability to generate text (§ 2.1), their potential to produce propagandistic and manipulative content (§ 2.2), and their pragmatic abilities compared to those of humans (§ 2.3).[4]

### 2.1. Human- and LLM-generated texts

The comparison between human and LLM-generated political speeches should be understood within the broader context of research on human and LLM-generated texts. Previous studies (e.g., Kreps et al., 2022; Muñoz-Ortiz et al., 2024) have demonstrated the ability of LLMs to mimic human language. In particular, human participants in several experiments were unable to distinguish between texts written by humans and those generated by LLMs. For instance, Kreps et al. (2022) found that participants in their experiment could not reliably differentiate between news articles written by humans and those produced by GPT-2. The high quality of LLM-generated writing has also been confirmed by studies involving domain experts. Casal & Kessler (2023) showed that academic linguists were generally unable to distinguish between LLM- and human-generated abstracts of research articles in Applied Linguistics. Similarly, Herbold et al. (2023) found that German high-school teachers consistently rated GPT-generated argumentative essays, particularly those created with GPT-4, higher than essays produced by students (L1 German speakers writing in English). However, in-depth linguistic analyses reveal significant differences between the two text sources. For example, examining register variation, Berber Sardinha (2024) observed that LLMs often over-rely on abstraction and struggle with referential language, especially in academic writing.

It is important to stress that these findings concern English, the language with the highest representation in LLM training datasets (see Brown et al., 2020: 14 for GPT-3). Comparable studies in less well-represented languages are still relatively scarce. Research on GPT-generated texts in Italian, for example, suggests that while their overall quality is high, they still contain unreliable information and exhibit problems with text organization that enable language experts to distinguish them from human-authored writing (Cicero, 2023). Moreover, the literature consistently reports evidence of English interference in AI-generated texts in less well-represented languages (for Italian see, for example, Cicero, 2023: 739; De Cesare, 2023; Tavosanis, 2024: 18-20).

### 2.2. Persuasion, manipulation, and LLMs: New tools for propaganda?

Several studies have highlighted the potential dangers that the widespread accessibility of LLMs pose to society. Goldstein et al. (2023) argue that LLMs could aid propagandists, since they 'could expand access to a greater number of actors, enable new tactics of influence, and make a campaign's messaging far more tailored and potentially effective' (Goldstein et al., 2023: 1). Similar concerns have also been raised by researchers concerned about microtargeting in advertising 'tailored to the recipient' (Burtell and Woodside, 2023: 3), particularly in the domain of political beliefs. Additional risks associated with LLMs include their ability to produce disinformation (Buchanan et al., 2021) and their susceptibility to political bias (Fujimoto and Takemoto, 2023; Liu et al., 2022). These concerns are further amplified by findings, discussed in § 2.1, that humans often struggle to distinguish between texts generated by LLMs and those written by humans.

However, some research has also emphasized the positive societal impact of LLMs. Several experiments have revealed that GPT models can effectively detect propaganda (Makhortykh et al., 2023; Kuznetsova et al., 2025). For instance, Kuznetsova et al. (2025) found that GPT-4 was successful at identifying fake news and texts referencing conspiracy theories in areas such as Covid-19 and the Holocaust. However, the results may also point to a possible language bias: while LLMs achieve human-like accuracy when detecting false information in English, their performance declines significantly when the material is in other languages, such as Russian and Ukrainian.

### 2.3. LLM and presuppositions

One of the most intriguing questions in LLM research concerns their ability to mimic aspects of human verbal communication. Recent studies, such as Hu et al. (2022) and Barattieri di San Pietro et al. (2023), found that larger models like Flan-T5 (Chung et al., 2022) and GPT-3.5 performed at near-human levels in many communication-oriented tasks but struggled with heavily context-dependent ones such as recognizing irony and certain types of metaphors. Barattieri di San Pietro et al. (2023) also noted GPT-3.5's issues with repetition and over-explaining in discourse tasks. However, Ruis et al. (2023) showed that fine-tuning techniques like *chain-of-thought* prompting can improve performance, with GPT-4 matching or exceeding human ability in interpreting conversational implicatures, as also suggested by Bojic et al. (2023) and Yue et al. (2024).

Several studies have also focused on presupposition (§ 2.4), a topic that lies at the heart of this paper and will therefore be reviewed in greater detail. Cong (2022), for example, designed a test in which transformer-based models such as DistillBERT and GPT-3 were asked to evaluate minimal pairs like (1a,b), drawn from a psycholinguistic experiment by Singh et al. (2016) on the accommodation of definite descriptions (§ 2.4.1) in more or less plausible contexts:

(1)

a. Seth went to jail. *The guard* spoke to him there for a while.
b. # Seth went to a restaurant. *The guard* spoke to him there for a while.

Since the presence of a guard is more natural in (1a) than in (1b), accommodating the presupposition that there was a (unique) guard in (1a) should be more natural, as confirmed by Singh et al.'s (2016) findings with human participants. However, GPT-3 performance was mostly at chance level, while DistillBERT performed somewhat better.

Kabbara & Cheung (2022) tested BERT and RoBERTa using the ImpPres dataset (Jeretič et al., 2020), designed to evaluate these models' ability to handle presuppositions and implicatures. They carried out two experiments, one of which involved modifying sentences from the ImpPres dataset by removing the presupposition trigger (§ 2.4.1). For example, in the minimal pairs in (2a, b), the definite description *Rose's drawing* in (2a) was removed in (2b). In the experiment, however, (2a, b) were both linked to (2c), which makes the presupposed content of *Rose's drawing* explicit.

(2)

a. Rose's drawing stunned the audience
b. Rose stunned the audience

[4] A *caveat* is necessary here. The literature on LLMs and their linguistic capacities spans a wide range of models, both older and more recent, as well as diverse linguistic phenomena, making it difficult to form a comprehensive view of the field.

c. Rose has a drawing

The results revealed that, on average, the BERT-based models tended to incorrectly recognize (2c) as a presupposition also for (2b). In the second experiment, the models often relied on superficial lexical and structural cues and their performance dropped significantly when specific words were removed. For instance, in a subset of the data containing cleft sentences like (3a) and their presupposition, in (3b), the models correctly recognized the contradiction between (4a) and (4b), which negates the cleft presupposition (someone talked about Cheryl), with 100 % accuracy:

(3)

a. It is Keith who stunned Christina
b. Someone stunned Christina

(4)

a. It is Helen that talked about Cheryl
b. No one talked about Cheryl

The models' accuracy dropped, however, when contradictory statements were presented without the negative pronoun *no one*, suggesting that the models relied on surface-level lexical cues to identify presupposition rather than a truly "pragmatic understanding of the presupposition at hand" (Kabbara and Cheung, 2022: 783).

Sravanthi et al. (2024) used the Pragmatic Understanding Benchmark (PUB) to evaluate some LLMs' ability to identify various pragmatic phenomena (implicature, presupposition, deixis and reference). Their results regarding presuppositions mirrored those of Kabbara & Cheung (2022): LLMs tended to rely on superficial linguistic forms instead of genuine pragmatic reasoning. Even the best-performing models, such as Llama-2–70B (Meta AI, 2023) and Flan-5, showed a considerable performance gap compared to the human evaluators.

Similar patterns were also observed in Sieker & Zarrieß (2023), who investigated the use of definite and indefinite determiners in English and German. Using BERT and several of its variants, they assessed models' predictions regarding the choice of determiners in unique, (5b), or non-unique, (5c), conditions following a context sentence, (5a):

(5)

a. Jan's mother was shopping. She bought one banana and two pears.
b. Of these, Jan received [*a* | *the*] banana (unique condition; correct choice: *the*)
c. Of these, Jan received [*a* | *the*] pear (non-unique condition; correct choice: *a*)

The results showed that, regardless of the conditions, LLMs consistently favored the indefinite determiner, even in cases such as (5b) where humans typically select the definite form, as shown in Schneider et al.'s (2019) findings. According to Sieker and Zarrieß (2023), this happened because the models were primed by lexical cues in the prompt (i.e., the presence of *one* in 5a).

Overall, the literature discussed in §§ 2.1–2.3 points to the following conclusions: (i) LLMs are capable of generating convincing persuasive texts and propaganda, in some cases achieving human-like performance, (ii) LLMs can also successfully detect fake news and propaganda; and (iii) some models outperform others in these tasks, with GPT-4 delivering the most promising results. However, despite these advances, (iv) LLMs still struggle, not unexpectedly, with some highly context-dependent phenomena. Furthermore, (v) findings on presupposition suggest that apparent successes in identifying inferential patterns may rely on superficial formal cues rather than proper pragmatic reasoning.

## *2.4. Presupposition*

### *2.4.1. Presupposition triggers*

In the field of philosophical and pragmatic studies (Karttunen, 1973; Sbisà, 2007; Stalnaker, 2002), presupposition has been described as one of the most pervasive and manipulative devices in human communication. The effects of its use in certain communicative contexts - such as political or commercial propaganda - have been very well documented in both theoretical and experimental lines of research (Lombardi Vallauri, 2009; Masia et al., 2017; Domaneschi et al., 2018). Since Stalnaker (1973; 2002), presuppositions have generally been characterized as information that is taken for granted by both the speaker and the receiver at the moment of utterance, i.e., content that is assumed by all interactants to be already known and true.

Their presence in a linguistic message is typically indexed by specific lexical categories or structures, therefore called presupposition triggers (Kiparsky and Kiparsky, 1971; Sbisà, 2007; Lombardi Vallauri, 2009, among others). Some triggers are specialized to presuppose the existence of certain referents (existence presuppositions), others presuppose the truth of certain states of affairs (truth presuppositions). Examples of some of the most common types of triggers are given in (6):

(6)
a. Definite description:
Demoliremo **l'Euromostro** (G. Meloni).
'We'll smash **the Euromonster**'
b. Change-of-state verbs:
**…commencer à bâtir** une Europe-puissance …' (E. Macron).
'**start building** a European power'
c. Iterative adverbs:
Make America great **again** (R. Reagan; D. Trump)

In (6a), the definite phrase *l'Euromostro* ('the Euromonster') presents the existence of a Euromonster as knowledge already shared by the addressee, when in fact it is a brand new lexical creation of the politician with the aim of increasing citizens' fear of the European Union and its policies. In (6b), the French verbs *commencer* ('to begin') and *bâtir* ('to build') convey the presuppositions that nobody has ever built a *Europe-puissance* ('European power') before, a piece of information that some - especially those opposed to Macron's views - might consider tendentious. In (6c), the iterative adverb *again* in Trump's well-known slogan presupposes that America was great in the past, but that this status has sometimes been weakened by previous administrations. Other relevant presupposition triggers discussed in the literature are adverbial subordinate clauses (*When John bought his new house*, his parents moved to the country), defining relative clauses (*The book I'm reading* is about Irish immigrants), factive predicates (*I'm surprised* that John decided to quit his job), among others (for a more exhaustive list, see Sbisà, 2007; Lombardi Vallauri, 2009, among others). One of the most striking properties of presuppositions in discourse is their resistance to negation; in other words, they persist even if the whole utterance is negated. For example in (7a,b),

(7)

a. It is not true that the black cat has crossed the street
b. It is not true that John has stopped smoking

only the asserted part of the sentences is negated (i.e., 'crossed the road' and 'stopped smoking', respectively), while the existence of a black cat and the fact that John used to smoke are still presupposed.

Resistance to negation is a clear logical consequence of the particular epistemic status presuppositions assign to some information and to the speaker themselves. In other words, by presupposing some content, the speaker presents that content as no longer negotiable and therefore to be accepted as it is for the purposes of the conversation. This property makes presupposition particularly effective in 'passing on' some meanings subliminally, that is, without the addressee being able to exercise the necessary epistemic vigilance (Sperber et al., 2010) to properly weigh up their truth (Lombardi Vallauri, 2019). Moreover, since presuppositions can easily be 'skipped' by our attentional frames - especially when relying on shallow processing mechanisms - their detection and interpretation and consequent reformulation as assertion is also less straightforward and requires ad hoc explanatory practices (Sbisà, 2021).

The ability of presuppositions to subtly instill content into the minds of recipients, often leading them to form potentially false beliefs, is what makes this strategy particularly manipulative, especially when used in certain contexts, such as political or commercial propaganda. As Rigotti (2005) rightly points out, its manipulative power lies in its function to create preferred mental models, often beneficial to the speaker but potentially harmful to the receiver.

#### 2.4.2. *Potentially manipulative presupposition*

As a strategy to present information as already part of the shared common ground in both oral and written linguistic messages, presupposition is often used to signal to the receiver that its content can be processed less thoroughly (Lombardi Vallauri, 2019), since it ostensibly does not add any new knowledge to the contextual set, in contrast, for example, to implicatures, which are generally associated with new information for the receiver (Sbisà, 2021: 179). Presupposition is thus, among other things, a highly economical means of avoiding the overt assertion of content that is not intended to fulfill the speaker's informative goal in the ongoing interaction. However, this particular discourse behavior also makes presupposition a good candidate for passing on potentially tendentious content without the addressee being able to question its truth (Lombardi Vallauri and Masia, 2014). Indeed, if a content has to be taken as already shared, its truth is neither assessed nor further weighed. For this reason, presupposition is considered to lead to a 'tacit' acceptance of its truth, since any attempt to question it would lead to an uncooperative conversational move (Sbisà, 2007). But what happens when the content in question is one that requires more attentive processing? If we take the context of political discourse, what happens when attacks, praise or opinions are conveyed by the politician in such a way as to be accepted without any room for contrast? Since we are mainly interested in the manipulative use of presuppositions, in our analysis we looked for such types of 'ideologically loaded' presuppositions, i.e., presuppositions that encode 'something that is questionable or even false and of which the source wants to convince the addressee(s)' (Lombardi Vallauri et al., 2020: 112). Presuppositions of this type, i.e., those that do not convey *bona fide* true content (Lombardi Vallauri, 2019) are referred to in this study as potentially manipulative presuppositions (henceforth, PMP).

#### 2.4.3. *Discourse functions of presuppositions*

Following an established tradition in prior studies (Cominetti et al., 2024; Garassino et al., 2022, 2024), we analyzed PMPs deductively, using the taxonomy of discourse functions shown in (8) (for convenience, presupposition triggers have been bold-typed):

(8)

a. CRITICISM (CRT): The content conveys a *criticism or an attack* against other political parties or politicians; moreover, there should be an *explicit target* to which this criticism or the attack is addressed.
Noi demoliremo **l'Euromostro** (G. Meloni).
'We'll smash the Euromonster'

b. SELF-PRAISE (SPR): Praise to personal achievements or accomplishments of their own party.
Nous **poursuivrons** sans relâche notre travail pour l'inclusion de nos compatriotes en situation de handicap (E. Macron).
'We will relentlessly continue our work for the inclusion of our compatriots with disabilities'

c. PRAISE OF OTHERS (OPR): Praise to other persons, also involving appreciation, condolences, wishes, etc.
Heureuse d'être ici, chez nous, sur **cette terre si française, populaire et patriote**, du bassin minier (M. Le Pen).
'I'm happy to be here, at home, in this land that is so French, so popular and so patriotic, in the mining district.'

d. STANCE-TAKING (STK): The content includes pieces of the political program and/or propaganda (e.g., political slogans); it may reflect values and ideological stances of the politician and/or their party.
Hanno cancellato **l'unico strumento di sostegno contro la povertà** (E. Schlein).
'They canceled the only tool to fight poverty.'

## 3. Research questions

Building on the foregoing, the study intends to address one overarching research question (ORQ):

ORQ. How closely do LLMs (specifically GPT-4) mimic the discourse of French and Italian politicians?

This question can be split into three further sub-questions:

RQ1. Is the frequency of PMPs and their presupposition triggers comparable between politicians' speeches and GPT-generated texts?
RQ2. Is the discourse function of PMPs in GPT texts similar to that in the politicians' speeches?
RQ3. Are there identifiable patterns in the use and distribution of PMPs that are characteristic of GPT-generated texts?

RQ1: The frequency of PMPs

Regarding the first sub-question, based on previous research (Garassino et al., 2024), we expect that PMPs are more frequent in GPT-generated texts. Moreover, it is also to be expected that the most common presupposition triggers in both politicians' and GPT's texts will be definite descriptions and change-of-state verbs.

RQ2: The discourse function of PMPs

Prior studies (Brocca et al., 2016; Garassino et al., 2022) have identified patterns regarding the role of the speaker in political discourse. Politicians in office tend to use PMPs with a stance-taking (STK) function, while opposition politicians tend to rely on PMPs to convey Criticism (CRT). More recently, Garassino et al. (2024) observed an additional trend: GPT-generated texts display a greater frequency of STK.

RQ3: Different patterns between LLMs and human politicians in the use of presuppositions

Previous research (Garassino et al., 2024) highlighted a strong association between GPT texts and the use of change-of-state verbs. In this paper, we aim to test whether this pattern holds in the GPT-4 output and, if so, to explore them more thoroughly through a qualitative lens to better understand how GPT-4 use of presuppositions may subtly differ from the way presuppositions are typically used by politicians.

In sum, while some aspects of the three research questions have been partially explored in Garassino et al. (2024), the use of a new, more powerful model (GPT-4) and a new dataset provide fresh opportunities for inquiry, reinforcing the relevance and novelty of these RQs.

## 4. Methodology

### 4.1. The corpus

The corpus examined in this study consists of two types of data, classified by source:[5] speeches delivered by politicians and texts generated by GPT-4. The political speech component includes two speeches each from French politicians Emmanuel Macron and Marine Le Pen, as well as Italian politicians Elly Schlein and Giorgia Meloni. These politicians were selected because they represent opposing ideological positions in their respective countries (see Table 1), and their speeches have been widely covered by the media. All speeches were delivered in public forums, such as rallies before the 2019 European elections, the 2022 French presidential election, or during party assemblies. The speeches by Italian politicians were delivered between 2018 and 2023, while those from French politicians were delivered between 2021 and 2022. At the time of their speeches, Emmanuel Macron was President of France, whereas Giorgia Meloni had not yet assumed the role of Prime Minister in Italy. In order to ensure thematic homogeneity, the central themes of these speeches were the European Union and immigration policy. For each politician, two speeches (T1 and T2) were selected, resulting in a total number of eight texts.

The GPT component of the corpus was generated using GPT-4, with the aim of creating a counterpart to each speech in the politicians' part (GPT_T1 and GPT_T2). To generate the text, the prompt in (9) was used:

(9)
'Imagine that you are [name of the politician], secretary of the [French/Italian] party [name of the party], delivering a speech at a rally in [place] during the [year] elections.

a. Continue the following speech: [first lines from a chosen politician's speech].
b. In your speech, address the [immigration/European Union] topic.
c. Create a speech in [French/Italian] consisting of approximately 4000 characters.'

In (9a), the prompt sets the context for the speech by following the *persona pattern* (Ekin, 2023; White et al., 2023) and aiming for specificity. This was done by including details such as the politician's name, their political party, the location and year of the speech to accurately replicate real-world conditions and create a realistic simulation. The prompt then provided a few selected lines from the politicians' speeches, (9b), directed the model to focus on a specific topic – such as immigration or the European Union – and specified both the language and the length of the speech in (9c).

All texts were generated in February 2024 by accessing the premium version GPT-4 on the ChatGPT website. We relied on the user-interface provided by ChatGPT, deliberately avoiding using the OpenAI API, in order to maintain a process that closely mirrors how most users interact with chatbots. Additionally, potential differences between outputs generated via API and the human-chatbot interface remain unexplored, as no systematic investigations on this issue have been conducted to date (Kuznetsova et al., 2025). All prompts were in English as previous studies (Haman, 2024; Lai et al., 2023) have indicated that GPT is optimized for tasks prompted in this language. We also used English for prompting in both French and Italian to avoid introducing different prompt language as a variable. Considering that the amount of training data for GPT in French and Italian is comparable – but substantially smaller than in English (Johnson et al. 2022; Lai et al., 2023) – we assumed that GPT's capability to produce outputs in both languages would also be comparable.

A total of sixteen texts were collected, eight of which were generated by GPT-4 and the remaining eight were collected from politicians. The speeches delivered by the politicians are longer than their GPT counterparts, a limitation resulting from technical restrictions on the model's output length. Following Garassino et al. (2024), who observed that LLM-generated texts tend to hallucinate when overly long, we limited the text length to ensure the generation of coherent texts.[6] Political speeches were trimmed from the end to match LLM output length.

The structure of the corpora and their lengths are presented in Tables 1–3.

It was assumed that the two sources could be considered broadly comparable in terms of diamesic variation. The human-produced texts in the corpus are not instances of spontaneous conversation, but rather monologic speeches that often follow pre-prepared scripts. Similarly, the GPT-generated texts share features with scripted speech, rather than reflecting spontaneous oral discourse (Labbé et al., 2024).

Finally, the variables extracted for the study are as follows: Discourse Function (CRT, OPR, SPR, STK), Language (French or Italian), Presupposition Trigger (CSV, DEF, and others; the complete list is available in the codebook on OSF, see footnote 5), and Source (GPT or Politicians).

### 4.2. Interrater reliability

The data were coded independently by Author 1 (DG) and Author 3 (VM). In the first annotation phase, the task was to identify the occurrences of PMPs in the two corpora. At this stage, no formal interrater reliability metrics were computed; instead, agreement was reached through negotiation (*consensus through negotiation*, see Loewen and Plonski, 2016: 90). Specifically, items labeled by both authors were automatically retained, while discrepancies were discussed. If consensus could not be reached, the item was excluded from the dataset. Once a shared set of PMP occurrences was obtained, both authors independently annotated the data according to the variables 'Presupposition Trigger' and 'Discourse Function'. Interrater reliability was then assessed using the Cohen's $k$ index (Table 4):

As expected, interrater reliability was considerably higher for the variable 'Presupposition Trigger' than for 'Discourse Function', the latter revealing the inherent subjectivity involved in the evaluation of pragmatic categories (see the discussions in Garassino et al., 2022, 2024). Despite the lack of consensus in the literature, Cohen's *k* values ranging between 0.61 and 0.80 are generally considered satisfactory (Landis and Koch, 1977: 165), which applies to most of our data (Table 4). Notable exceptions are the 'Discourse Function' annotations for Schlein's and Macron's texts, which yielded lower *k* values, indicating only moderate agreement. To address this issue, a second round of *consensus through negotiation* was conducted: non-agreed cases were jointly reviewed by the authors and, as in the identification phase of PMPs, any instance on which agreement could not be reached was excluded.

**Table 1**
Politicians selected for the corpus.

| French | Italian |
|---|---|
| Emmanuel Macron<br>*La République en Marche!*<br>President of the French Republic<br>(Center) | Elly Schlein<br>*Partito Democratico*<br>(Left) |
| Marine Le Pen<br>*Rassemblement National*<br>(Right) | Giorgia Meloni<br>*Fratelli d'Italia*<br>(Right) |

**Table 2**
Length of the politicians' speeches.

| | Characters T1 | Characters T2 | Characters Tot |
|---|---|---|---|
| Le Pen | 5508 | 5155 | 10,663 |
| Macron | 5731 | 5330 | 11,061 |
| Meloni | 5758 | 5267 | 11,025 |
| Schlein | 3338 | 7010 | 10,348 |

[5] All texts, the codebook, annotated data, and R scripts are publicly available on the OSF website: https://doi.org/10.17605/OSF.IO/A65CV

[6] In one instance (Meloni, T2), GPT exceeded the prompt's request by generating a text of 4,219 characters. The lengths of the generated texts vary across cases, and no interventions were made to modify GPT's output, in order to preserve the ecological validity of the study.

**Table 3**
Length of the texts generated by GPT-4.

| | Characters T1 | Characters T2 | Characters Tot |
|---|---|---|---|
| Le Pen | 3842 | 3425 | 7267 |
| Macron | 3697 | 3794 | 7491 |
| Meloni | 3717 | 4219 | 7936 |
| Schlein | 3522 | 3337 | 6859 |

**Table 4**
Interrater agreement indices (Politicians and their GPT versions).

| | Politicians | | GPT | |
|---|---|---|---|---|
| | Presupposition Trigger | Discourse Function | Presupposition Trigger | Discourse Function |
| | *k* | *k* | *k* | *k* |
| Le Pen | 0.97 | 0.66 | 0.82 | 0.68 |
| Macron | 0.89 | 0.43 | 0.84 | 0.65 |
| Meloni | 0.96 | 0.78 | 0.91 | 0.68 |
| Schlein | 0.94 | 0.43 | 0.86 | 0.65 |

All in all, this process resulted in a final dataset consisting only of fully agreed-upon occurrences.

## 5. Analysis

### 5.1. *Frequency, form and function*

Due to the limited size of the dataset, this section relies on data visualization and descriptive statistics. In relation to RQ1, the absolute frequency of PMPs in the corpus is shown in Table 5:

Although a cursory glance at Table 5 suggests that the frequency of PMPs is higher in the politicians' speeches, the normalized frequencies presented in Figs. 1 (aggregated data) and 2[7] (individual data by politician and corresponding chatbot version) indicate that the frequency of PMPs is on average slightly higher in GPT-generated texts than in the politicians' speeches (the only exception being Schlein).

The distribution of the presupposition triggers is presented in Fig. 3.

**Table 5**
Absolute frequencies of PMPs ($N = 278$).

| Politicians | PMPs | GPT | PMPs |
|---|---|---|---|
| Le Pen | 57 | Le Pen_GPT | 39 |
| Macron | 31 | Macron_GPT | 23 |
| Meloni | 31 | Meloni_GPT | 31 |
| Schlein | 42 | Schlein_GPT | 24 |
| **Total** | **161** | | **117** |

Notably, change-of-state verbs (CSV) and definite descriptions (DEF) account for over 80 % of occurrences in the GPT data and approximately 65 % in the politicians' speeches. All other triggers represent only a small fraction of the data individually.

Regarding RQ2, the discourse functions most frequently associated with PMPs are stance-taking (STK) in the GPT data, and criticism (CRT) and stance-taking (STK) in the politicians' data, as illustrated in Fig. 4:

An analysis of individual variation reveals greater functional variety in the politicians' data compared to GPT, where stance-taking emerges as the only consistently widespread discourse function of PMPs, as shown in Fig. 5.

These findings mirror those reported in Garassino et al. (2024: 13–15), which focused on GPT-3.5. In that study, the overall frequency of PMPs was likewise higher in GPT-generated texts. Garassino et al. (2024) also observed the dominance of CSV and DEF as presupposition triggers in GPT-3.5 data, along with their divergent distribution in LLM-generated versus politician-authored speeches. Similarly, STK emerged as the most frequent discourse function of PMPs in the GPT data, while CRT also featured prominently in the politicians' texts. Notably, in Garassino et al. (2024), CRT was fairly frequent in the GPT-3.5 dataset, contrasting with the current findings, in which criticism appears more limited in the GPT-4 data (except for Le Pen_GPT).

Finally, to summarize the results observed in Figs. 1–5, we present heatmaps (Fig. 6) that focus exclusively on the most frequent presupposition triggers, CSV and DEF.[8]

The color gradient in Fig. 6 directly reflects the magnitude of counts for each tile. Examining the darker tiles in the GPT data, we can observe that most PMP examples are represented by CSV with a STK discourse function, regardless of the language, as in the examples in (10). In contrast, in the politicians' dataset, many PMPs are represented by DEF with a CRT function, as in the examples in (11):

(10)
a. Nous devons **bâtir une véritable politique de défense européenne, investir dans nos capacités militaires**, et **renforcer notre autonomie stratégique** (Macron_GPT, T2)
'We must build a true European defense policy, invest in our military capabilities, and strengthen our strategic autonomy.'
b. In quest'ottica, intendiamo lavorare per **rafforzare i canali di migrazione legale** (Schlein_GPT, T1)
'With this in mind, we intend to work to strengthen legal migration channels.'

(11)
a. Le voir débattre victorieusement, le voir affronter et mettre en déroute les représentants de **l'oligarchie méprisante, la triste et prétentieuse Madame Loiseau**, ça a été pour moi, et j'en suis sûre aussi pour vous, un immense motif de satisfaction (Le Pen, T2)
'To see him debate victoriously, to see him confront and defeat the representatives of the contemptuous oligarchy, the sad and pretentious Madame Loiseau, that was for me, and I'm sure for you too, an immense source of satisfaction.'
b. Questo accade a causa delle vostre scelte e **del vostro decreto inumano**, che ha il solo scopo di rendere più difficile salvare le vite in mare (Schlein, T1)
'This happens because of your choices and your inhumane decree, which serves only to make it harder to save lives at sea.'

In (10a), *bâtir* ("to build") presupposes that there has never been a true European defense policy; *investir* ("to invest") presents as already known that nobody has ever invested in the country's military capabilities, and *renforcer* ("to strengthen") conveys the presupposition that the country's strategic autonomy used to be weaker. In the Italian occurrence in (10b), *rafforzare* ("to strengthen") encodes as already known that legal immigration channels were weaker. As for the examples in (11), the French one in (11a) contains the presuppositions that there is a contemptuous oligarchy and that Madame Loiseau is sad and pretentious. By the same token, the definite phrase *del vostro decreto inumano* ("of your inhuman decree") presupposes the existence of an inhuman decree.

[7] The figures were created with R (v. 4.4.3; R Core Team, 2025) and the ggplot2 package (v. 3.5.1; Wickham, 2016).

[8] The presupposition triggers labeled with REL (defining relative clauses) were counted along with DEF, if the nominal head was definite. If the nominal head was indefinite, the occurrence was not counted. The dataset represented in Fig. 6 is therefore smaller (N = 211) than the dataset on which Figs. 1–5 are based (N = 278).

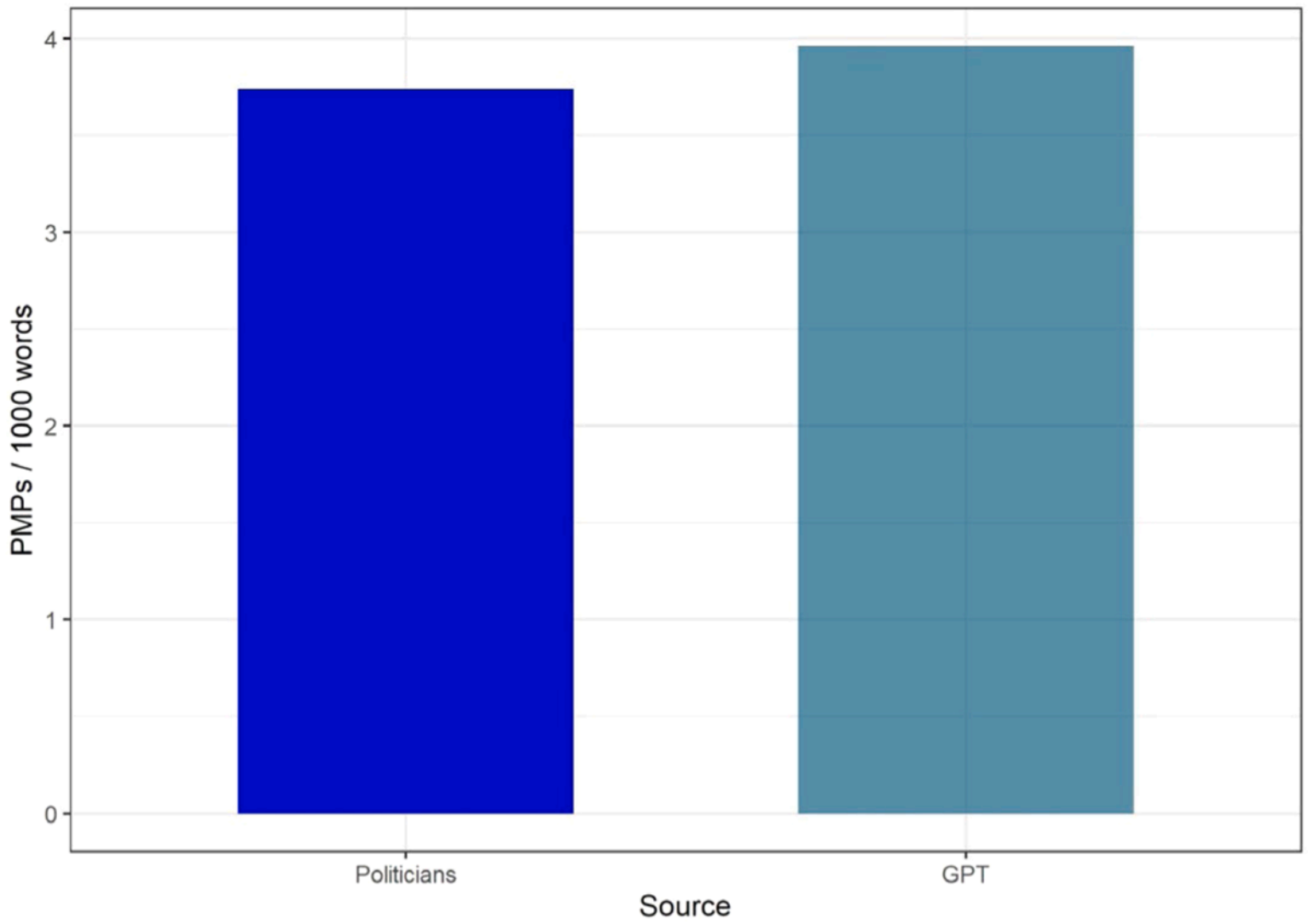

**Fig. 1.** Normalized frequencies of PMPs (per 1000 words).

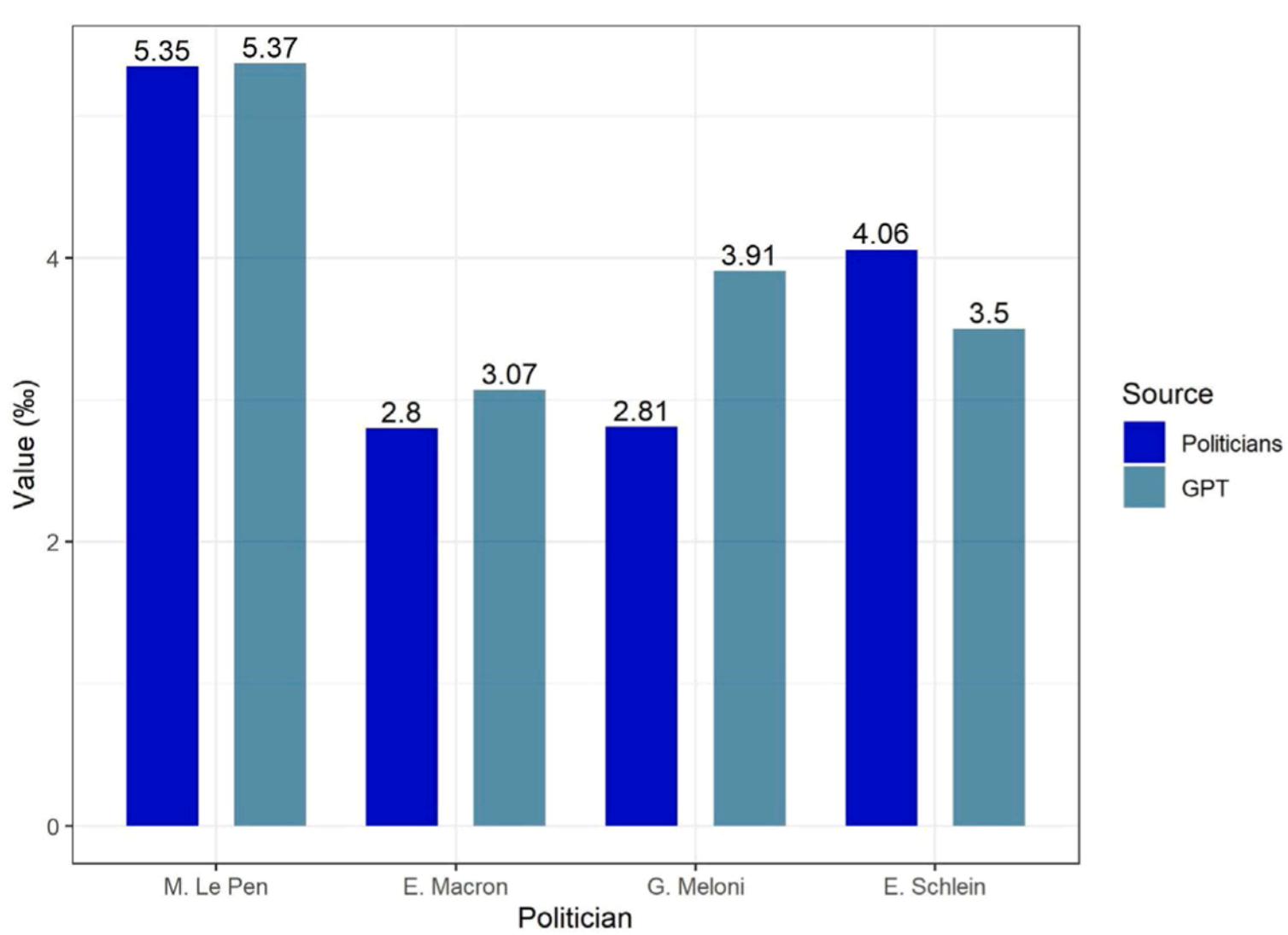

**Fig. 2.** Normalized frequencies of PMPs (per 1000 words).

Finally, some differences between French and Italian politicians emerge. For instance, the French politicians in the corpus use DEF in a more varied way than their Italian counterparts, employing it not only to convey CRT but also functions such as OPR (12a) and SPR (12b):

(12)
a. **Cette page d'histoire que vous avez écrite ici,** dans la cité d'Hénin, ici, dans le bassin minier au cœur du pays ouvrier, n'est plus seulement votre histoire mais, soyez-en honorés, un peu d'histoire de la France et, dimanche prochain, de l'Europe (Le Pen, T2)
'This page of history that you have written here, in the city of Hénin, here in the mining basin at the heart of the working-class country, is no longer just your story but, be proud, a part of France's history and, next Sunday, of Europe's as well.'
b. Et je veux vous dire ce soir que je mettrai toutes mes forces pour convaincre chacun que **le seul procès pour le pouvoir d'achat**, c'est le nôtre ! Que **le seul projet crédible contre la vie chère**, c'est le nôtre ! (Macron, T2)
'And I want to tell you tonight that I will put all my energy into convincing everyone that the only real plan for purchasing power is ours! That the only credible project to fight the high cost of living is ours!'

Interestingly, the data examined in this section suggest that there may be a special 'affinity' between GPT data and the STK function expressed by CSV, indicating a distinct pattern that sets LLM-generated texts apart from those produced by politicians. In § 5.2, we will further examine the occurrence of CSV in GPT data.

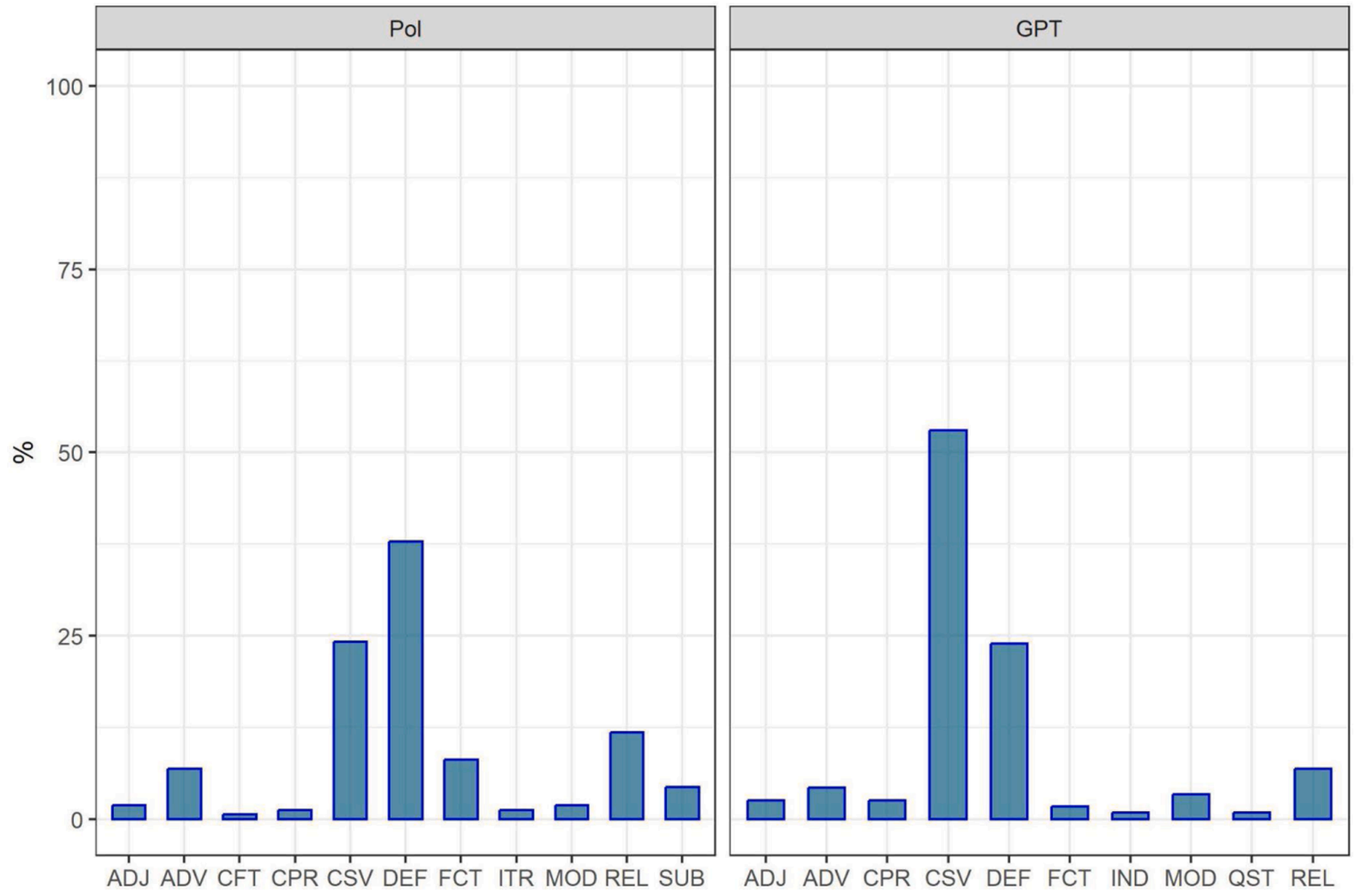

**Fig. 3.** Distribution of presupposition triggers (percentage) in the GPT-4 and the politicians' data (ADJ= adjective, ADV = adverb, CFT = cleft sentence, CPR = comparative, CSV= change-of-state verb, DEF = definite description, FCT = factive verb, IND = indefinite description, ITR = iterative verb, MOD = modal verb, QST = question, REL = defining relative, SUB = adverbial subordinate).

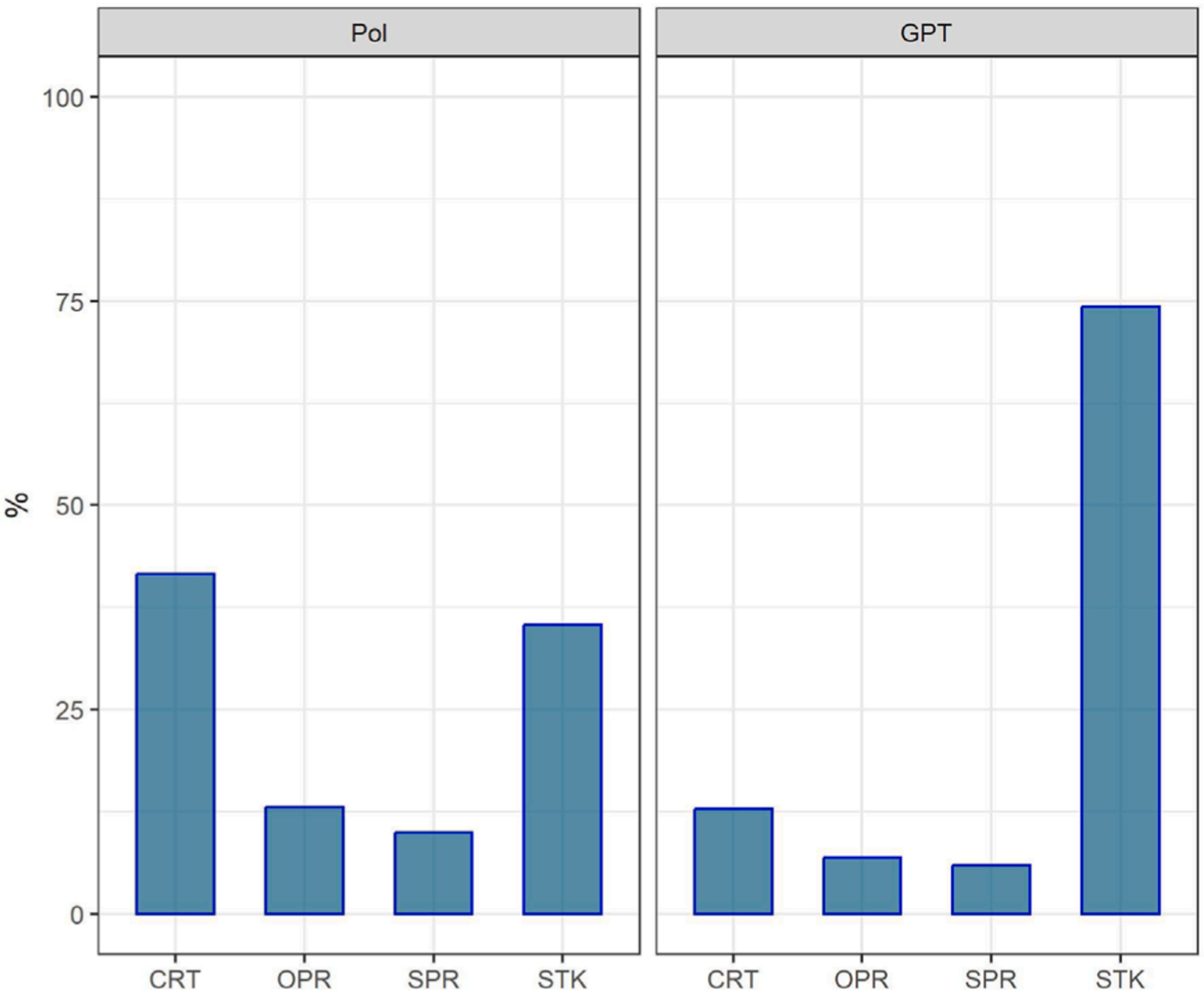

**Fig. 4.** Distribution of discourse functions (percentage) in the GPT-4 and the politicians' data (CRT = criticism, OPR = praise of others, SPR = self-praise, STK = stance-taking).

### *5.2. LLM-specific patterns: change-of-state verbs*

In the light of the predominance of CSV as presupposition triggers in GPT data, as shown in § 5.1, a more fine-grained analysis is required. Six common change-of-state verbs were selected from the corpus (each verb having an equivalent in the other language). The verb pairs selected were: *défendre/difendere*, 'to defend', *protéger/proteggere*, 'to protect', *bâtir/costruire*, 'to build', *renforcer/rinforzare*, 'to reinforce', and *favoriser/promuovere*, 'to promote'. Although data-driven, this analysis was also informed by previous findings in the literature: specifically, Labbé et al. (2024) observed an overuse of certain verbs, most notably the change-of-state verb *continuer* ('to continue') in GPT-generated speeches compared to those delivered by French presidents.

The data were divided into eight sub-corpora according to sources (human vs GPT) and politicians (Le Pen, Macron, Meloni, Schlein). The sub-corpora were uploaded into the Sketch Engine software (Kilgarriff et al., 2014). Each subcorpus was examined for the presence of the selected CSV conveying PMPs. The frequency of the selected CSVs was compared between politicians and their GPT-generated counterparts. The absolute and relative frequencies of CSVs were higher in the texts generated by GPT, as demonstrated in Table 6. For instance, in the GPT corpus of Schlein, the number of selected CSVs was 17, in contrast to the unique occurrence observed in the human counterpart. Upon normalization to the total corpus length, the former is shown to correspond to 2.4 additional occurrences per thousand words in the GPT-generated corpus in comparison to the human-authored speeches. In contrast, for Macron, the GPT-generated texts contain 13 occurrences, while his speeches contain 10 occurrences, resulting in a difference of 0.8 occurrences per thousand words.

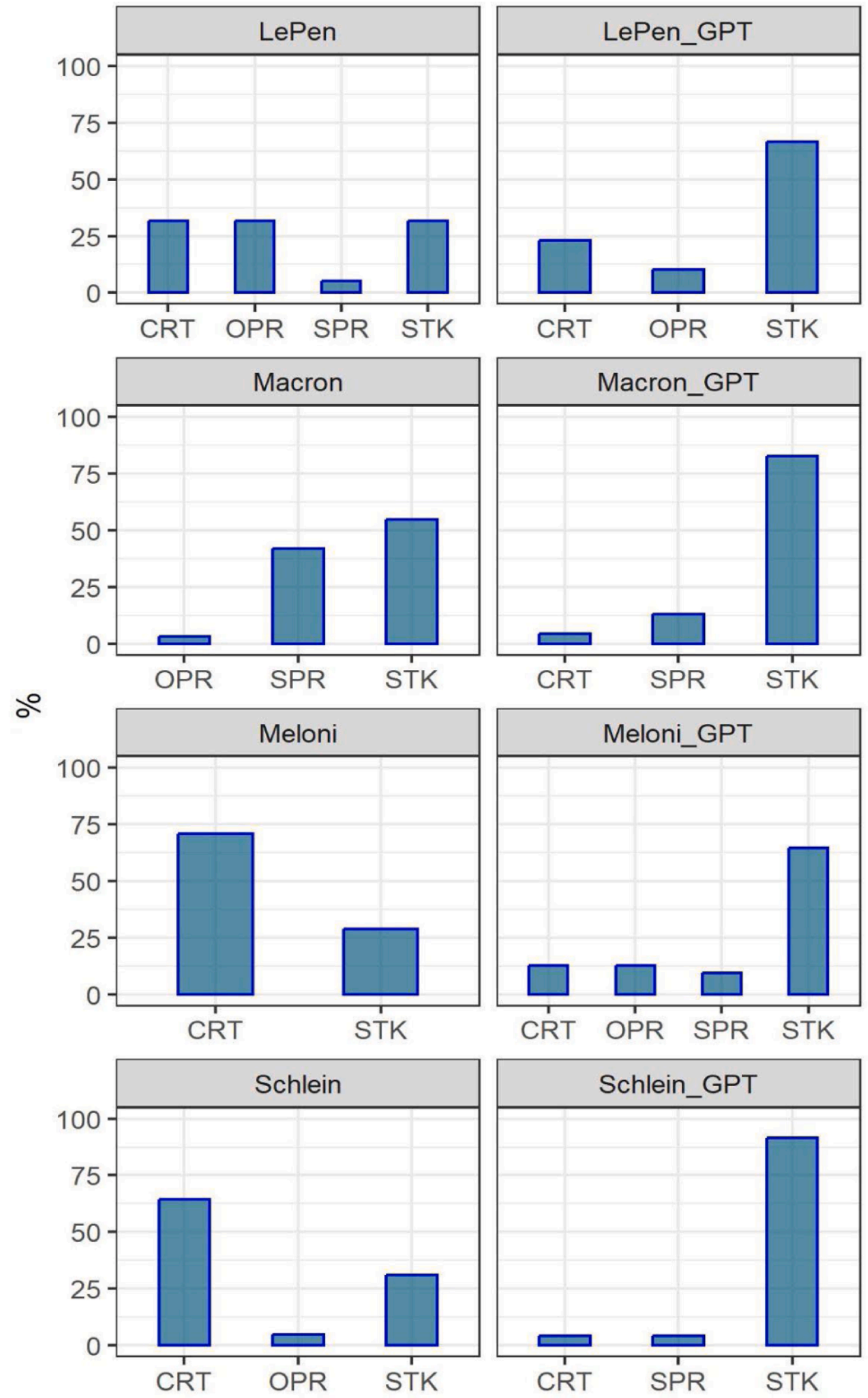

**Fig. 5.** Distribution of discourse functions (percentage). Focus on individual variation.

From a more qualitative perspective, the contexts in which the selected CSVs are used were also analyzed. This research identified association patterns between the selected verbs and their respective subjects or direct object arguments, revealing a tendency for these verbs to co-occur with a limited set of arguments that reflect the ideological and political orientations of individual politicians. In the case of Macron, (13a), the semantic pattern involving 'to defend' and its object is linked to economic values, whereas for Meloni, (13b), and Le Pen, (13c), it relates to relevant issues in the conservative agenda, such as immigration, national values, and safety. Finally, for Schlein, (13d), this pattern aligns with typical left-wing topics, such as social rights and multiculturalism:

(13)
a. Une Europe […] capable de **défendre ses intérêts économiques, culturels et stratégiques**. (GPT_Macron, T2)
'A Europe […] capable of defending its economic, cultural, and strategic interests'
b. Insieme, possiamo […] **difendere i nostri valori** e lavorare per un'Italia più sicura. (GPT_Meloni, T1)
'Together, we can […] defend our values and work for a safer Italy'
c. Une Europe […] qui **défende ses frontières.** (GPT_Le Pen, T1)
'A Europe that defends its borders'
d. …**difendiamo un'Europa aperta, multiculturale, che rispetti e valorizzi le diversità**… (GPT_Schlein, T2)
'…we defend an open, multicultural Europe that respects and values diversity'

Conversely, the occurrence of the verb 'to defend' in the corpus of politicians does not exhibit clear patterns, and its collocations appear more variable. This variability may contrast with the GPT-generated texts, where this verb seems to recur in more consistent combinations, possibly reflecting the model's attempt to align with the style of the mimicked politician. However, given the relatively low frequency of the verb, such observations should be interpreted with caution.

In certain examples of GPT data, similar arguments are shared by politicians with opposing ideological views, yet the modifiers reflect more closely their political orientations. For example, in (14) the pattern involving the verb 'to build' with the direct object argument 'Europe' occurs in the data of three different politicians, but the context and the adjectives or the defining relative clauses modifying the nominal heads are suggestive of their respective political orientations:

(14)
a. Nous devons **bâtir une Europe qui écoute et qui s'adapte aux aspirations de ses peuples**. (GPT_Le Pen, T1)
'We must build a Europe that listens to and adapts to the aspirations of its peoples'
b. Pour **bâtir une France et une Europe plus fortes, plus accueillantes et plus unies**. (GPT_Macron, T1)
'To build a stronger, more welcoming, and more united France and Europe'
c. Possiamo **costruire un'Italia e un'Europa più giuste, più umane, più forti**. (GPT_Schlein, T1)
'We can build a fairer, more humane, and stronger Italy and Europe'

These examples highlight how LLMs adapt common association patterns to suit the ideological diversity of different politicians. These findings are crucial as they demonstrate that while GPT-generated texts can successfully mimic rhetorical patterns typical of political discourse, they tend to do so through formulaic and uniform structures.

## 6. Discussion

We can now come back to the three research questions that have been addressed in this paper:

RQ1. Is the frequency of PMPs and their presupposition triggers comparable between politicians' speeches and GPT-generated texts?
RQ2. Is the discourse function of PMPs in GPT texts similar to that in the politicians' speeches?
RQ3. Are there identifiable patterns in the use and distribution of PMPs that are characteristic of GPT-generated texts?

With regard to RQ1, the overall frequency of PMPs is on average slightly higher in GPT-generated texts than in politicians' speeches.

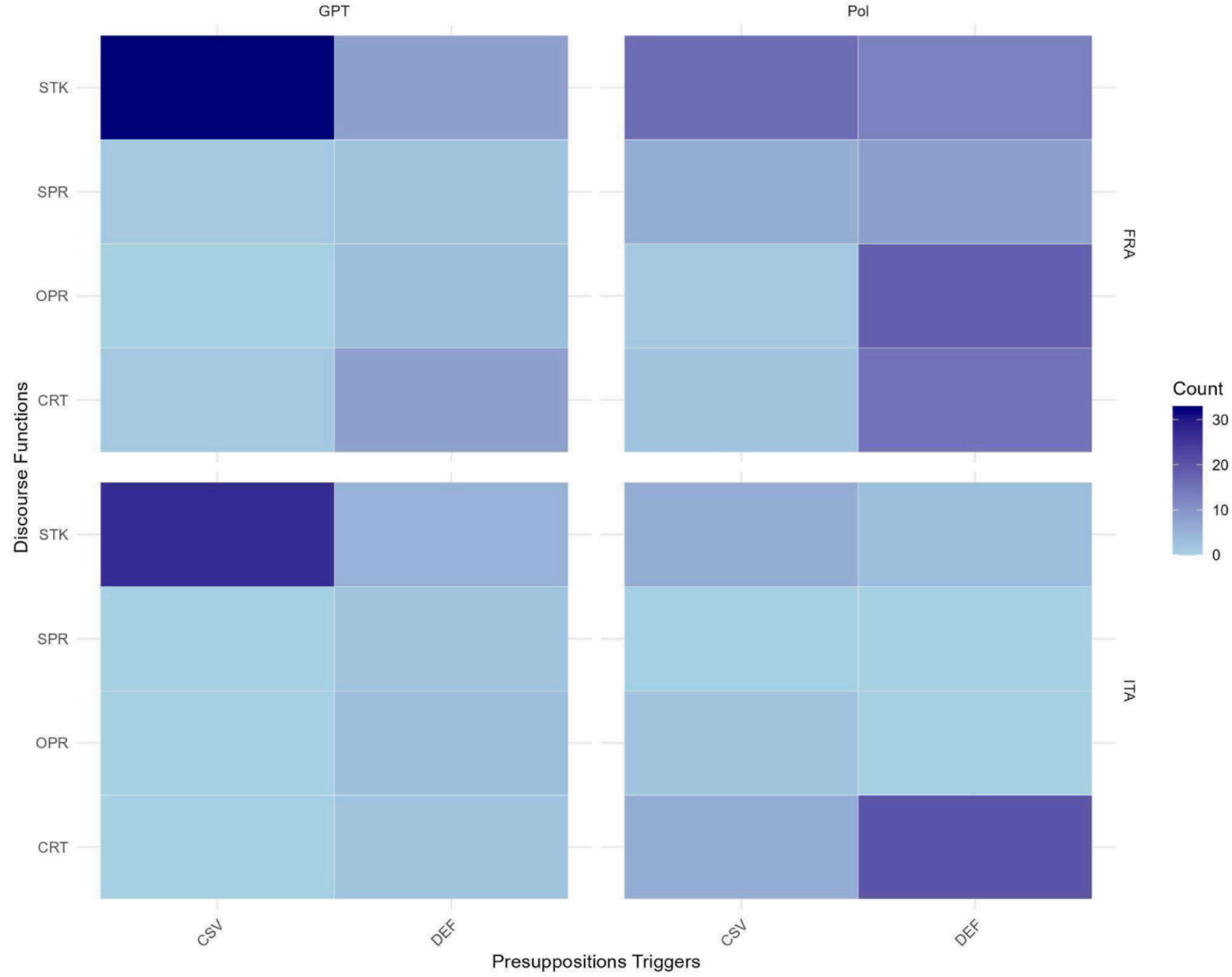

**Fig. 6.** Heatmaps showing the relationship between the variables ‘Presupposition Trigger’, ‘Discourse Function’, ‘Language’ and ‘Source ’.

**Table 6**
Occurrence of selected CSVs (in French and Italian) across each subcorpus.

| | Le Pen | | Macron | | Meloni | | Schlein | |
|---|---|---|---|---|---|---|---|---|
| | GPT | Pol | GPT | Pol | GPT | Pol | GPT | Pol |
| *défendre/difendere* | 2 | 1 | 5 | 6 | 2 | 2 | 3 | 1 |
| *protéger/ proteggere* | 4 | 0 | 2 | 0 | 2 | 0 | 2 | 0 |
| *bâtir/costruire* | 3 | 0 | 4 | 4 | 2 | 2 | 5 | 0 |
| *renforcer/ rinforzare* | 2 | 1 | 2 | 0 | 1 | 0 | 2 | 0 |
| *favoriser/promuovere* | 0 | 0 | 0 | 0 | 1 | 0 | 5 | 0 |
| **Sum** | **11** | **2** | **13** | **10** | **8** | **4** | **17** | **1** |
| **Frequency in the subcorpus** | **1.5 ‰** | **0.4 ‰** | **1.7 ‰** | **0.9 ‰** | **0.9 ‰** | **0.3 ‰** | **2.5 ‰** | **0.1 ‰** |

Additionally, notable differences emerge in the distribution of presupposition triggers: DEF is more frequent in the politicians’ data, while CSV is more prevalent in the GPT data.

Concerning RQ2, the discourse functions of PMPs also differ markedly in the two sources. In GPT-generated texts, PMPs are predominantly used to express STK, whereas in politicians’ speeches, their functions are less homogenous, with CRT and STK emerging as the most frequent.

As for RQ3, the paper examined the distribution of CSVs in the corpus. This analysis revealed recurring patterns in GPT texts, typically involving verb phrases with a CSV as the head followed by various object nouns. These nouns vary depending on the politician being emulated and may be accompanied by connotated modifiers. For example, a left-leaning politician is more likely to use phrases such as ‘defend an open Europe,’ while a right-leaning politician might express the need to ‘defend the borders.’ Such recurring combinations are less frequent in the politicians’ speeches, where a wider variety of verb-object pairings reflects greater linguistic flexibility, which LLMs may not fully capture.

This contrast suggests a core feature of how LLMs operate: driven by statistical predictions, they tend to reproduce n-grams, such as verb-object combinations, typical of a specific discourse genre, adapting them semantically to match the politician’s profile in question. However, in doing so, LLMs also appear to amplify certain patterns typical of political speech, leading to their over-representation.

From a broader perspective, these findings, together with those by Kabbara & Cheung (2022), Sieker and Zarrieß (2023) and Sravanthi et al. (2024), suggest that LLMs tend to rely on superficial cues when processing inferential phenomena. This tendency appears to be language-independent in GPT data, as evidenced by the similar patterns observed in both French and Italian. In other words, the frequent use of PMPs in LLM-generated texts may be due to the high salience of certain recurrent word combinations in political discourse. As a result, PMPs in LLM-generated texts tend to cluster around CSV predicates predominantly associated with a STK function. This distribution differs from that found in politicians’ speeches (see Fig. 5 and 6), where PMPs display a broader functional spectrum.

Despite these differences, LLMs such as GPT-4 are nevertheless capable of replicating potentially manipulative linguistic strategies, including the use of presuppositions that convey not *bona fide* content.

## 7. Conclusion

This study has shown that, although politicians’ and GPT-generated texts addressing the same discourse topics may appear superficially very similar, they differ significantly in their pragmatic characteristics. In particular, the distribution of presuppositions in GPT data appears to be due to the statistical prominence of n-grams that involve CSVs. Moreover, while presuppositions in politicians' speeches serve a wider range of discourse functions, GPT-generated texts tend to employ them predominantly for STK, resulting in a more limited functional spectrum.

However, several limitations of the present inquiry must be acknowledged. First, some of them stem from the intrinsic characteristics of LLMs. Our decision to use GPT-4 in an 'ecological setting' directly via the web interface did not allow us to control or track parameters such as temperature. This limitation, combined with the continuous evolution of LLMs, may render research findings quickly outdated and difficult to replicate, posing significant challenges for longitudinal studies and reproducibility. Second, limitations may arise from the prompting strategy employed. While we relied on a *zero-shot* strategy (see Brown et al., 2020), recent research suggests that other strategies, such as *chain-of-thought* prompting, can yield higher-quality outputs across different tasks (e.g., Brocca et al., in press). Third, the overall amount of data analyzed, both from politicians' speeches and GPT-generated texts, remains limited. This limitation stems from the labor-intensive process of annotating pragmatic categories, which must be carried out manually. Reducing the workload involved in such annotation, potentially with the assistance of AI, remains a key desideratum (Brocca et al., under review).

Despite these constraints, our research suggests notable differences between human and GPT-generated political speeches, particularly regarding the distribution and content of PMPs. However, these differences are often subtle and difficult to detect without specialized tools, underscoring the potential risks LLMs pose in shaping political and social discourse.

Given LLMs' tendency to rely on potentially manipulative language, some educational applications can be envisioned. One promising direction involves the development of didactic programs aimed at enhancing both digital literacy skills and users' ability to make implicit content explicit, as proposed by Brocca et al. (2024). As Sbisà (2021) convincingly argues, the explicitation of implicit meaning is a necessary step to maximize the understanding of a text, enabling critical engagement - namely, the evaluation or questioning of the truth conveyed by a linguistic message. Beyond educational implications, research into how LLMs detect manipulative language, particularly potentially manipulative presuppositions, can inform strategies for fine-tuning these models through more focused and precise prompts (Sieker et al., 2024).

## Declaration of generative AI and AI-assisted technologies in the writing process

During the preparation of this work the authors did not use any generative AI or AI-assisted technologies in the writing process. The authors used AI-assisted technologies (GPT-4 and DeepL) for the linguistic revision of the paper.

Supplementary material associated with this article can be found at https://doi.org/10.17605/OSF.IO/A65CV.

## CRediT authorship contribution statement

**Davide Garassino:** Writing – original draft, Methodology, Data curation, Writing – review & editing, Resources, Formal analysis, Visualization, Investigation, Conceptualization. **Nicola Brocca:** Resources, Writing – review & editing, Investigation, Writing – original draft, Data curation. **Viviana Masia:** Writing – review & editing, Formal analysis, Writing – original draft, Investigation.

## Declaration of competing interest

The authors declare that they have no known competing financial interests or personal relationships that could have appeared to influence the work reported in this paper.